\documentclass[10pt]{article}
\usepackage[margin=2cm]{geometry}
\usepackage{lipsum}
\usepackage{amsfonts}
\usepackage{graphicx}
\usepackage{booktabs,arydshln}
\usepackage{amsmath,mathrsfs}
\usepackage{float}
\usepackage{xcolor}
\title{\textbf{Interpretable Multimodal Emotion Recognition using Facial Features and Physiological Signals}}
\author
{
    Puneet Kumar and Xiaobai Li\footnote{Corresponding Author: xiaobai.li@oulu.fi}\\
    CMVS, University of Oulu, Finland.\\
    \small\textit\{puneet.kumar, xiaobai.li\}@oulu.fi\\
}
\date{}
\begin{document}

%%%%%%%%%%%%%%%%%%%%%%%%%%%%%%%%%%%%%%%%%%%%%%%%%%%%%%%%%%%%%%%%%%%%%%%%%%%%%%%%%%%%%%%%%%%%%%%%%%
\maketitle
\begin{abstract}
This paper aims to demonstrate the importance and feasibility of fusing multimodal information for emotion recognition. It introduces a multimodal framework for emotion understanding by fusing the information from visual facial features and rPPG signals extracted from the input videos. An interpretability technique based on permutation feature importance analysis has also been implemented to compute the contributions of rPPG and visual modalities toward classifying a given input video into a particular emotion class. The experiments on IEMOCAP dataset demonstrate that the emotion classification performance improves by combining the complementary information from multiple modalities.\vspace{.03in}

\end{abstract}

%%%%%%%%%%%%%%%%%%%%%%%%%%%%%%%%%%%%%%%%%%%%%%%%%%%%%%%%%%%%%%%%%%%%%%%%%%%%%%%%%%%%%%%%%%%%%%%%%%
\section{Introduction}
Emotions, characterized by a rich and complex mix of physiological and cognitive states, hold significant importance across multiple fields such as psychology, human-computer interaction, affective computing, and even extending to broader domains such as virtual reality, user experience design, healthcare, and education~\cite{poria2017review}. Understanding and accurately interpreting emotions is essential in human communication and social interactions~\cite{cimtay2020cross}. With the surge in the development and accessibility of multimodal sensing technologies, researchers can explore multiple modalities to enhance the accuracy and robustness of emotion recognition systems~\cite{baltruvsaitis2018multimodal}. The current research trend focuses on building Artificial Intelligence (AI) systems that can be deployed for real-life applications~\cite{paleyes2022challenges}.\vspace{.03in}

Two such modalities, facial expressions and physiological signals, have garnered significant attention due to the rich information they offer and their non-invasive nature~\cite{yu2021facial}. Facial expressions, direct and non-invasive indicators of emotion, have been thoroughly investigated~\cite{malik2021towards}. Various techniques involving the extraction of facial landmarks, local descriptors, or holistic representations have been proposed to capture nuanced variations in facial muscle movements that reflect different emotional states~\cite{wang2018facial}. Physiological signals, such as remote photoplethysmography (rPPG) signals, provide another layer of emotional cues. These signals, obtained through non-contact video-based techniques, offer insights into physiological changes associated with emotional responses~\cite{yu2021facial}. The interplay of these two modalities offers a more holistic understanding of emotions, thus enhancing the robustness of emotion recognition systems~\cite{zeng2009survey}.\vspace{.03in}

Emotion classification through audio-visual information is a well-established research task~\cite{rao2019learning,xu2020improve,majumder2019dialoguernn}. However, recognizing emotion using the physiological context along with the audio-visual information score for further exploration~\cite{yu2021facial}. Furthermore, despite the significant advancements, many multimodal emotion recognition models do not provide meaningful interpretations for their predictions~\cite{murdoch2019definitions,longo2020explainable}. Most existing interpretability techniques have been implemented for visual modality and have yet to be fully explored for multimodal analysis~\cite{ribeiro2016should,selvaraju2017grad,malik2021towards}.\vspace{.03in}

This paper proposes an interpretable multimodal emotion recognition framework that extracts rPPG signals and facial features from the input videos and uses their combined context for emotion detection. The Haar cascades classifier ~\cite{soo2014object} has been implemented to extract the rPPG signals, whereas a pre-trained ResNet-34-based network extracts the visual features. Further, early and late fusion approaches that integrate the static facial expression features and dynamic rPPG signals to capture both spatial and temporal aspects of emotions have been incorporated.\vspace{.03in}

An interpretability technique based on permutation feature importance (PFI)~\cite{altmann2010permutation} has also been incorporated that computes the contribution of rPPG and visual modality towards classifying a given input video into a particular emotion class. The experiments performed on Interactive Emotional Dyadic Motion Capture (IEMOCAP) dataset~\cite{busso2008iemocap} have resulted in an accuracy of 54.61\% while classifying the input videos into ten emotion classes (`neutral,' `happy,' `sad,' `angry,' `excited,' `frustrated,' `fearful,' `surprised,' `distressed' and `other'). The increased performance on using the multimodal context than the individual accuracies on using rPPG or visual modality alone advocates the importance of leveraging the multimodal context for emotion understanding. The average contributions of rPPG and visual modalities towards emotion recognition have been computed as 37.67\% and 62.33\%, respectively.\vspace{.05in}

\noindent The contributions of this paper can be summarized as follows:\vspace{-.09in}
\begin{itemize}
\item A multimodal emotion recognition framework has been proposed to classify a given video into discrete emotion classes. It extracts the dynamic rPPG signals from the input videos and combines them with static facial expressions using early and late fusion approaches.\vspace{-.1in}
\item An interpretability technique has been incorporated that computes the contribution of rPPG and visual modalities towards emotion classification using the PFI algorithm.\vspace{-.1in}
\item Extensive experiments have been performed on the IEMOCAP dataset, and the results have been presented in terms of accuracy, precision, recall, F1 score, and modality-wise contributions toward emotion classification.
\end{itemize}

%%%%%%%%%%%%%%%%%%%%%%%%%%%%%%%%%%%%%%%%%%%%%%%%%%%%%%%%%%%%%%
\section{Proposed Method}
\begin{figure}[h]
\centering
\includegraphics[width=.62\textwidth]{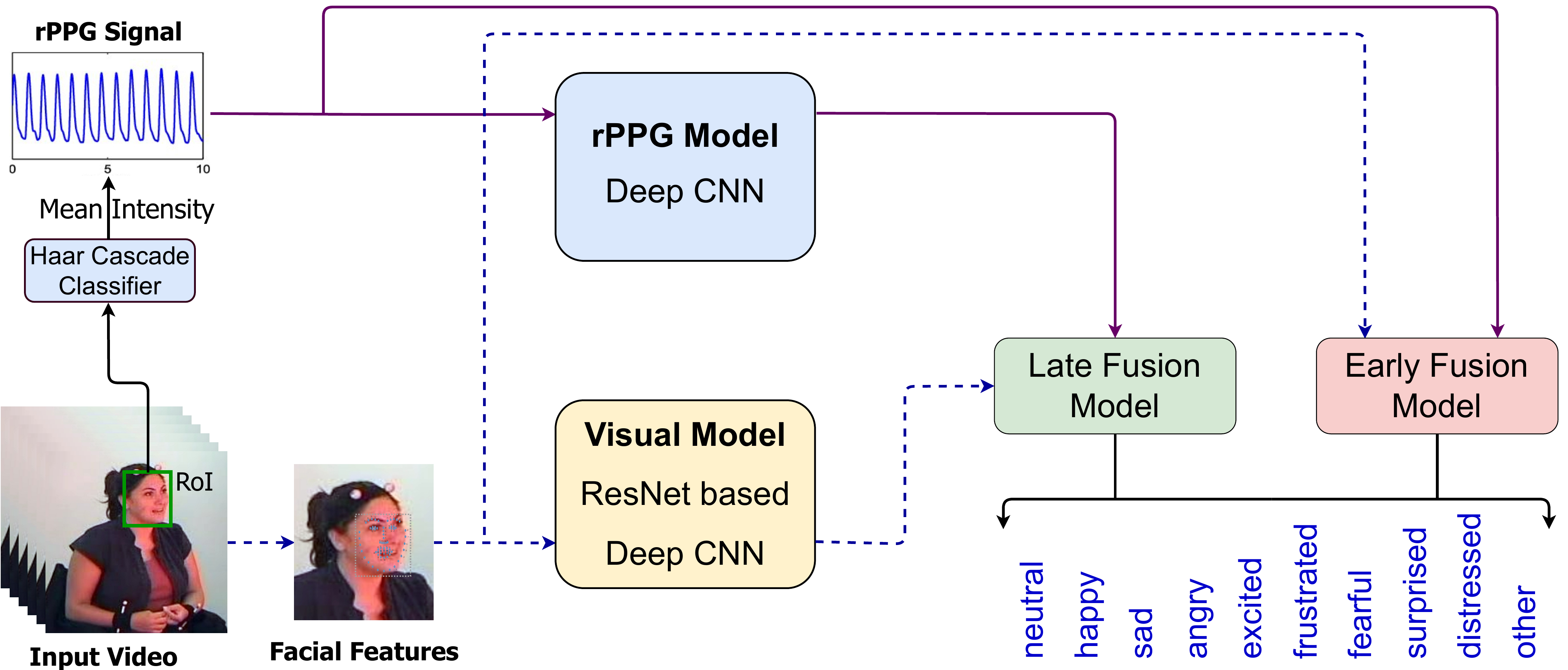}
\caption{Schematic illustration of the proposed framework.}
\label{fig:archi}
\end{figure}

The proposed framework has been diagrammatically depicted in Figure \ref{fig:archi} and described in the following sections. 

\vspace{-.15in}

\subsection{Preprocessing and Feature Extraction}
The video files are loaded and processed frame by frame using OpenCV (cv2) library \footnote{https://opencv.org/} and processed to extract rPPG signals and facial features.\vspace{.07in}

% Mean intensity extraction from a region of interest (ROI) is the facial area detected using the Haar cascade classifier. Specifically, the code takes the mean of the green channel (indicated by frame[y:y+h, x:x+w, 1]) for each detected face in each frame of a video. It then applies a detrending operation to remove any linear trends.
\noindent\textit{i) rPPG Signals Extraction}: Face detection within each video frame during the rPPG signal extraction process is accomplished using Haar cascades~\cite{soo2014object}. The region of interest (ROI), predominantly the facial region, is isolated from each frame, after which the mean intensity is computed to generate the rPPG signal for each video. The calculation of the mean intensity within the ROI ($\bar{I}c$) is represented in Eq. \ref{eq:1}.\vspace{-.1in}

\begin{equation}
\bar{I}c = \frac{1}{N} \sum_{x=1}^{W}\sum_{y=1}^{H} I_{x, y, c}
\label{eq:1}
\end{equation}

Where $I_{x, y, c}$ is the intensity of the pixel at location $(x, y)$ for color channel $c$ in the ROI, and $N$ is the total number of pixels in the ROI, whereas $W$ and $H$ represent the width and height of the ROI, respectively, and $c \in {R, G, B}$.\vspace{.07in}

\noindent\textit{ii) Facial Features Extraction}: Facial feature extraction employs Dlib's shape predictor \cite{dlib2016davis}, which is a version of the ResNet-34 trained on Face Scrub dataset\cite{ng2014data} to identify the facial landmarks in a given image of a face. As per Eq. \ref{eq:2}, it identifies 68 facial landmarks for each detected face within every frame, distinguishing unique facial characteristics.\vspace{-.28in}

\begin{equation}
\begin{split}
    P &= D(F, \{L_i\}) \\
    F &= [f_1, f_2, \ldots, f_n]
\end{split}
\label{eq:2}
\end{equation}

Where $F$ represents the face detected in a frame, $P$ represents the predicted points on the face, $D(F, \{L_i\})$ is the function for predicting points on the face, and $L_i$ is the set of landmark points for the $i^{th}$ point. As signals from different videos might differ in length, it becomes crucial to standardize the input for the neural network model. This standardization is achieved by zero-padding $\bar{I}$ and $P$ to match the maximum signal length.

\subsection{Multimodal Feature Fusion}
Early fusion and late fusion approaches are used to combine the rPPG signals and facial features.\vspace{.07in}

\noindent\textit{i) Early Fusion}:
In the early fusion approach, the rPPG signals and facial features are concatenated before being fed into the model. The fused data are then passed through a neural network comprising a flatten layer, followed by CNN layers of dimensions 512 and 256, and the final layer of size equal to the number of classes. The flatten layer transforms the 3D input tensor into a 1D tensor, and the subsequent CNN layers functions perform the classification task. The model structure is represented as per Eq. \ref{eq:3}. 

\begin{equation}
\begin{aligned}
I' &= \text{concatenate}(\bar{I}c, P) \\
I'' &= \text{flatten}(I') \\
F_{early} &= \text{NNet}(I'', C) \\
\end{aligned}
\label{eq:3}
\end{equation}

Where $I$ is the input shape, $C$ denotes the number of classes, $\bar{I}c$ is the mean intensity within the ROI from the rPPG signals, $P$ represents the facial features, $NNet$ represents the early fusion network and $F_{early}$ is the output of the early fusion.\vspace{.07in}

\noindent\textit{ii) Late Fusion}:
In the late fusion approach, the rPPG and visual models are trained separately, and their outputs are combined using a weighted average. Eq. \ref{eq:4} represents a late fusion approach where the models are trained separately, and their outputs are combined in the final output $F_{late}$.\vspace{-.15in}

\begin{equation}
\begin{split}
F_{late} &= w_1 \cdot M_{\text{rPPG}}(\bar{I}c) + w_2 \cdot M_{\text{facial}}(P) 
\end{split}
\label{eq:4}
\end{equation}

Where $M_{\text{rPPG}}(\bar{I}c)$ and $M_{\text{facial}}(P)$ represent the outputs of the rPPG model and the visual model, respectively, and $w_1$ and $w_2$ are the weights assigned to each model's output in the final fusion.

\subsection{Emotion Classification}
This study employs three separate models for emotion classification. Two of these models operate independently, utilizing rPPG signals and facial features. The third model operates via `early fusion,' exploiting the combined context of data from the rPPG and visual models. The outputs of these individual models are then collaboratively integrated through a `late fusion' approach that uses a weighted addition technique. The individual models, based on rPPG signals and facial features, are constructed as follows.\vspace{.07in}

\noindent\textit{i) rPPG Model:} This model utilizes a Deep Convolutional Neural Network (CNN) with two hidden layers. It incorporates Rectified Linear Unit (ReLU) activation functions for emotion classification derived from rPPG signals.\vspace{.07in}

\noindent\textit{ii) Visual Model:} This model, built on facial features, employs a ResNet-based Deep CNN with two hidden layers and ReLU activation functions.

\subsection{Interpretability}
An explainability method based on permutation feature importance (PFI) \cite{altmann2010permutation} is implemented, which is used to estimate the importance of features by permuting the values of each feature and measuring the resulting impact on model performance. The PFI of feature $j$ is the decrease in the model score when values of feature $j$ are randomly permuted. PFI for a feature $j$ is the difference in the model score when the values of feature $j$ are randomly permuted. Eq. \ref{eq:pfi} mathematically represents the concept of permutation feature importance.\vspace{-.05in}

\begin{equation}
PFI(j) = E_{\pi}[f(X^{(i)})] - E_{\pi}[f(X^{(i)}_{\pi_j})]
\label{eq:pfi}
\end{equation}

Where $PFI(j)$ is the permutation feature importance of feature $j$, $E_{\pi}[f(X^{(i)})]$ is the expected value of the model score over all samples in the dataset when the model is scored normally, $E_{\pi}[f(X^{(i)}_{\pi_j})]$ is the expected value of the model score when the values of feature $j$ are permuted according to some permutation $\pi$, and $X^{(i)}_{\pi_j}$ denotes the dataset $X^{(i)}$ with the values of feature $j$ permuted according to $\pi$.

%%%%%%%%%%%%%%%%%%%%%%%%%%%%%%%%%%%%%%%%%%%%%%%%%%%%%%%%%%%%%%%%%%%%%%%%%%%%%%%%%%%%%%%%%%%%%%%%%%
\section{Results and Discussion}
\subsection{Experimental Setup}
The emotion classification experiments have been performed on the IEMOCAP dataset~\cite{busso2008iemocap} consisting of 10,039 videos labeled with ten discrete emotion labels (`neutral,'` happy,' `sad,'` angry,' `excited,' `frustrated,' `fearful,' `surprised,' `distressed' and`other'). The model training has been trained on NVIDIA RTX 4090 GPU for 50 epochs with a batch size of 32 and a learning rate of 0.001. The performance has been evaluated using accuracy, precision, recall, and F1 score metrics.

\subsection{Results}
Table \ref{tab:resultsDetailed} summarizes the accuracy of the individual and fusion models, whereas the average contributions of rPPG and visual modalities towards emotion recognition in the early fusion setup are presented in Table \ref{tab:contribution}. The proposed framework has demonstrated an emotion classification accuracy of 54.61\%, and the average contributions of rPPG and visual modalities towards emotion recognition have been computed as 37.67\% and 62.33\%, respectively.\vspace{-.07in}

\begin{table}[h]
\centering
\caption{Detailed performance of the individual and fusion models.}
\vspace{.033in}
\label{tab:resultsDetailed}
\begin{tabular}{ccccc}
\toprule
\textbf{Model} & \textbf{Accuracy} & \textbf{Precision} & \textbf{Recall} & \textbf{F1 Score} \\\midrule%\hline
rPPG & 37.45\% & 0.37 & 0.38 & 0.38 \\
Facial Features & 46.42\% & 0.49 & 0.49 & 0.49 \\
Late Fusion & 41.17\% & 0.43 & 0.42 & 0.42 \\
%\hdashline 
Early Fusion & 54.61\% & 0.56 & 0.58 & 0.57 \\
\bottomrule
\end{tabular}
\end{table}\vspace{-.2in}

\begin{table}[h]
\centering
\caption{Average contribution of each modality towards emotion recognition.}
\vspace{.033in}
\label{tab:contribution}
\begin{tabular}{cccc}
\toprule
\textbf{Modality} & \textbf{Contribution}\\\midrule %\hline
rPPG & 37.67\% \\
Visual & 62.33\% \\
\bottomrule
\end{tabular}
\end{table}

Table \ref{tab:resultsDetailed} shows that both the individual models performed reasonably well. However, the fusion model outperformed the individual models, demonstrating the advantage of combining rPPG signals and facial feature information for emotion recognition.

\subsection{Discussion}
%The findings emphasize the potential of combining visual and physiological cues, reinforcing the importance of multimodal context in emotion recognition.
This paper presents a compelling case for including multimodal context in emotion recognition. While the models trained on individual modalities show moderate performance, their fusion significantly improves emotion recognition accuracy. It emphasizes the complementarity of these modalities in capturing emotional states. However, the late fusion of modalities underperforms compared to the early fusion approach, indicating that integrating modalities at an earlier stage allows for more effective learning of emotional states. 

However, this study has a few limitations of the proposed work. The IEMOCAP dataset, while widely used, may limit the generalizability of the findings. Cross-dataset experiments on larger and more diverse datasets could further strengthen the results. Moreover, more modalities such as audio, text, and other physiological signals can also be incorporated for emotion recognition. Finally, a more in-depth interpretability mechanism can be developed to explain the role of individual features in emotion detection. 

\section{Conclusion}
This work presents a multimodal emotion recognition framework using rPPG signals and facial features. It paves the way for practical applications where transparent and interpretable emotion understanding is important. The results highlight the benefits of integrating multiple modalities for emotion recognition, with an early fusion approach yielding the highest accuracy. While there are limitations and potential improvements, our study provides a promising direction for future research in emotion recognition, emphasizing the importance of multimodal data and fusion techniques. %As the field advances, the development of robust, reliable, and efficient emotion recognition systems holds great potential in numerous applications, from mental health monitoring to human-computer interaction. %In the future, we will focus on addressing these limitations by conducting more extensive evaluations, investigating strategies for handling class imbalance, and further improving the generalization capabilities of the models. 

%%%%%%%%%%%%%%%%%%%%%%%%%%%%%%%%%%%%%%%%%%%%%%%%%%%%%%%%%%%%%%%%%%%%%%%%%%%%%%%%%%%%%%%%%%%%%%%%%%
\bibliographystyle{unsrt}
\bibliography{main}
\end{document}